\documentclass[final]{clv2025} 
\jvol{vv}
\jnum{nn}
\jyear{2025}

\usepackage{amsmath}
\usepackage{booktabs}
\usepackage{multirow} 

\usepackage{linguex}

\begin{document}

\runningtitle{}

\runningauthor{Proietti, Lenci}


\pageonefooter{Under Review}

\title{The \emph{quasi}-semantic competence of LLMs: \\ A case study on the \textit{part-whole} relation}



\author{Mattia Proietti\thanks{Corresponding author}$^{}$, Alessandro Lenci$^{*}$}

\affilblock{
    \affil{University of Pisa, CoLing Lab, Dept. of Philology, Literature and Linguistics\\\quad \email{mattia.proietti@phd.unipi.it }, \email{alessandro.lenci@unipi.it}}
}

\maketitle

\begin{abstract}
  Understanding the extent and depth of the semantic competence of \emph{Large Language Models} (LLMs) is at the center of the current scientific agenda in Artificial Intelligence (AI) and Computational Linguistics (CL). We contribute to this endeavor by investigating their knowledge of the \emph{part-whole} relation, a.k.a. \emph{meronymy}, which plays a crucial role in lexical organization, but it is  significantly understudied. We used data from ConceptNet relations \citep{speer2016conceptnet} and human-generated semantic feature norms \citep{McRae:2005} to explore the abilities of LLMs to deal with \textit{part-whole} relations. We employed several methods based on three levels of analysis: i.) \textbf{behavioral} testing via prompting, where we directly queried the models on their knowledge of meronymy, ii.) sentence \textbf{probability} scoring, where we tested models' abilities to discriminate correct (real) and incorrect (asymmetric counterfactual) \textit{part-whole} relations, and iii.) \textbf{concept representation} analysis in vector space, where we proved the linear organization of the \textit{part-whole} concept in the embedding and unembedding spaces. These analyses present a complex picture that reveals that the LLMs' knowledge of this relation is only partial. They have just a ``\emph{quasi}-semantic'' competence and still fall short of capturing deep inferential properties.
  
\end{abstract}

\section{Introduction}

Large Language Models (LLMs) have come lately to dominate the Natural Language Processing (NLP) landscape with their remarkable capabilities, their high performances on a wide range of tasks and the so-called ``emergent abilities'' \citep{wei2022emergent} they are allegedly able to acquire along with conspicuous brute force size increments \citep{kaplan2020scaling}. Despite the day-to-day improvements in generating text and addressing several NLP tasks emphasized by researchers and industrial practitioners, at a more theoretical level a lack of consensus still exists whether they have truly language understanding abilities \citep{Mitchell_2023-debate} and a full-fledged model of meaning \citep{piantadosi2022meaning, Pavlick2022}, whether they are capable of high-level generalizations and symbol manipulation and grounding \citep{Pavlick2023}, or they are simply very powerful probabilistic machines that have learned to mimic human behavior from the huge training data \citep{bender-koller-2020-climbing,Bender2021}. 
A point often raised by scholars is their struggle to cope with reasoning, factual and common sense knowledge, but also the degree of abstraction they can attain relatively to high-level concepts and semantic relations \citep{lenci:2023b,mahowald-dissoc-lang-tought}. This raises the issue of the true nature and scope of the \textbf{semantic competence} of LLMs.

In this work, we investigate one particular and yet crucial aspect of LLMs' inferential competence \cite{Marconi2003-nu}: their knowledge of the \emph{part-whole} relation (aka \emph{meronymy}) and their understanding of its core property of being \emph{antisymmetric} (see below Section \ref{pw-properties}). Rigorous analysis of the peculiar competencies of LLMs is a challenging tasks and requires careful task and dataset design.  A widespread method consists in \textbf{prompt-based behavioral analysis}, which when performed in comparison with the investigation of human behavior may play a key role in crafting special diagnostics to reveal useful information regarding both human and machine understanding. Nowadays models, exponentially increasing in size at a seemingly constant pace, are becoming more and more complex and it is becoming even harder to assess whether their improvements are due to data and architecture scaling, leveraging billions of parameters to learn statistical correlations from trillions of tokens, or to the genuine acquisition of higher-level cognitive capabilities. 
However, prompt-based behavioral tasks suffer several shortcomings and do not always align with the computations happening inside models, nor do they shed light on their internal representations of concepts and knowledge. Therefore, behavioral investigations must be complemented with other methods to get a more comprehensive insight into the nature of the models' competence.
For example, analyzing model responses in probability space can help get more faithful judgments of the models' abilities \citep{hu-levy-2023-prompting}. On the other hand, inspecting the internal representations yielded by pre-training may allow us to get a better understanding of whether and how concepts are encoded inside the models' semantic space \citep{park2024linearrepresentationhypothesisgeometry}.

We studied the LLMs' competence of meronymy adopting a comprehensive approach that consists of three levels of analysis:
\begin{enumerate}[i.)]
\item \textbf{behavioral level} -- 
we prompted the models about their declarative knowledge of the \textit{part-whole} relation to solve tasks consisting in answering questions, judging statements, generating parts for given objects, and recognizing their own generated parts; 
\item \textbf{probabilistic level} -- we judged the LLM's capabilities to assign higher scores to correct meronymic sentences (\emph{The wheel is a part of a the car}) rather than to their swapped version (\emph{The car is a part of a wheel}) violating the antisymmetry of the relation;

\item \textbf{representational level} -- we used the \textit{linear representation hypothesis}, recently adopted for the semantic mechanistic interpretability of LLMs \citep{Templeton:2024} and their internal geometrical representation of concepts \citep{park2024linearrepresentationhypothesisgeometry,Park2024TheGO}, to analyse the representation of the \textit{part-whole} relation in the vector spaces of their \textit{embedding} and \textit{unembedding} layers.
\end{enumerate}

\noindent{}The major findings from these analyses can be summarized as follows:

\begin{enumerate}
    \item models do not possess a very strong generalisation about the \textit{part-whole} relation and its core property of being \textit{antisymmetric};
    \item instruction-tuned models are far better at generating meronyms for given holonyms, than understanding the relation between them when asked;
    \item approaching the same problem from different methodological perspectives can yield mixed results about models' capabilities. This is particularly true when confronting analyses at the behavioral  (prompting) and probabilistic levels;
    \item the \emph{part-whole} concept in \textit{embedding} space is only partially recoverable by means of the Linear Representation Hypothesis;
    \item although class-specific parts are encoded in coherent subspaces, the embeddings of LLMs do not seem to encode a general \emph{part-whole} relation.
\end{enumerate}

\noindent{}All in all, our investigation shows that, despite LLMs' knowledge of meronymy is \emph{prima facie} impressive, they have just a partial approximation of this relation. A substantial gap with respect to human knowledge of meaning still exists in LLMs: They have just a \textbf{``\emph{quasi}-semantic'' competence}. 

This paper is organized as follows:  in Section \ref{part-whole-rel} we briefly introduce the part-whole relation and its relevant logic properties; in Section \ref{sec:exp-overview} we present an overview of the methodologies we apply along with the related works; in Section \ref{sec:data} we introduce the data sources we used and the process we followed to derive the testing benchmarks; in Section \ref{sec:models} we describe the LLMs we used for the experiments; 
in Sections \ref{behav-task-desc}, \ref{sec:sent-plaus} and \ref{linhypo-method} we details the methods, show and discuss the results of respectively our behavioral analysis, sentence plausibility modelling and analysis of the \textit{part-whole} concept in vector space; we then move to Section \ref{sec:wrap-up} for a wrapped up discussions of the results and limitations of the work and to Section \ref{sec:conc} to conclude.

\section{The part-whole Relation} \label{part-whole-rel}
The \emph{part-whole} relation is a fundamental structuring and organizing principle of entities in the world. Diverse disciplines and domains have endeavored to define it precisely and understand its nature and properties.

In philosophy, the enquiry about the nature of such a relation, also known as \emph{mereology} \citep{sep-mereology} seeks to establish a coherent ontological theory of the essence and the interaction between things and their parts, as well as a description of their status among the realm of world entities. 
In linguistics, among lexical-semantics relations, \emph{meronymy} plays a crucial role in the description of word meaning and the hierarchical organization of the lexicon \citep{Lyons1977, cruse-lexsem}. In NLP and Artificial Intelligence \emph{meronymy} is also a core component in the construction of knowledge bases and ontologies such as WordNet \citep{wordnet} and ConceptNet \citep{speer2016conceptnet}.

\subsection{What is a \textit{part}?} \label{sec:part notion}
Notwithstanding its importance, it is not easy to find a unified definition accounting for all the possible nuances of the \textit{part-whole} relation both due to its intrinsic nature and to its resemblance and affinity with other relations (e.g., groups and pieces).

The problems in defining meronymy relate to the difficulty of finding a satisfactory notion of \textit{part}. For instance, a \textit{part} has to be distinguished from a \textit{piece}, but still they are affine and share a number of commonalities. Indeed, they both involve the relation between a whole and some smaller entity which is integrated with it at some point in time. Following \citet{cruse-lexsem}, it is easy to acknowledge the distinction between what are the \textit{parts} composing an object and what are the \textit{pieces} obtained by cutting that same object into smaller entities. Cruse considers three distinctive properties of parts:  
\begin{itemize}
    \item \textit{autonomy}, that is the possibility of clearly identify the part as an entity of its own when separated from the whole;
    \item \textit{non-arbitrary boundaries}, which states that a "\textit{part is normally delimited from its sister parts by a relative discontinuity of some sorts}" \citep[p. 159]{cruse-lexsem};
    \item \textit{determinate funtion}, following which a part must serve a precise purpose in the economy of the whole it belongs to.
\end{itemize}

However, such properties in turn raise further issues. Consider for example an entity such a \textit{slice of cake}. Given that a slice is an arbitrary section of the whole cake, it would be considered as a \textit{piece} rather than a \textit{part} following the definitions above. However, other authors describe the notion of \textit{part} in terms of different features, which also apply to the case of cake slice. For example, \citet{Winston1987-taxo} define a \textit{part} as being \textit{functional} and/or \textit{separable} and/or \textit{homeomerouos} with respect to the whole. According to this view, not all the features of meronymy need to be present at the same time, allowing for a certain degree of gradience and variation. 
While the properties of being \textit{functional} and \textit{separable} may be alternatives to the definitions given above, respectively of \textit{determinate function} and \textit{autonomy},
\textit{homeomerous} means that a whole can be made of parts that are similar to each other and to the whole, with no clear-cut boundaries and delimitations.
This stance would account for cases like a \textit{sliced cake}, because it is still a cake made of its slices. 

Other problems arise in identifying linguistic phrases able to unequivocally express the part-whole relation. The two most used linguistic diagnostics to identify meronymy \citep{cruse-lexsem, cruse-transit, Lyons1977} are phrases like  
\textit{The X is a part of the Y} or \textit{The Y has a/an X} as shown in \ref{ex:mer-diag}:

\ex. \label{ex:mer-diag}
\a. \textit{The wheel is a part of the car}.
\b. \textit{The car has wheels/a wheel}.

However, these phrases are very polysemous and can express other semantic relations than the \textit{part-whole} one as happens in \ref{ex:counterex}, where neither sentence is referring to a meronymy relation:

\ex.\label{ex:counterex}
\a. \textit{Mary has a son}.
\b. \textit{The dog is part of the mammals}.

This contributes to make it difficult to find a non-ambiguous way to linguistically express the notion of \textit{part} and \textit{part of} and suggests to consider \textit{part} just as an umbrella term covering several types of relations that would be better represented using other linguistic expressions (i.e., \emph{component of, member of} etc.). 

Some authors have attempted to define detailed taxonomies of the possible meanings of \emph{part} \citep[see also below Section \ref{pw-properties}]{Winston1987-taxo, chaffin-emp-taxon}. One might for example want to distinguish \textit{necessary} from \textit{optional} parts in the first place. For example, consider a footstool and a cushion as parts of a sofa. It is not necessary for a sofa to have a footstool, but it is a part of the sofa if present. On the contrary, we can hardly have a sofa without a cushion. Another binary distinction can be drawn between \textit{segmental} and \textit{systemic} parts \citep{cruse-lexsem}, changing the axis along which we consider the \textit{holonym}. Segmental parts are spatially delimited and are typically encountered sequentially, as the rooms of a house or the limbs of a body. On the other hand, systemic parts are integrated and diffused through the whole \textit{holonym}, as the electric cables in a house or the veins in a body. All this is a cause, along with cognitive proximity, for the possible confusion between \textit{meronymy} and other close-related, sometimes overlapping, relations, such as the \textit{topological inclusion} one (e.g., \emph{The prisoner is in the cell}; \citealt{cruse-lexsem, Winston1987-taxo}).

\subsection{The properties of the part-whole relation} \label{pw-properties}
In the literature, the \textit{part-whole} relation is described as having the properties of being \emph{reflexive, transitive} and \emph{antisymmetric} \citep{sep-mereology}. 

\textit{Reflexivity} is a simple statement of identity, meaning that every entity is part of itself. Considering \textit{P} an expression of the meronymic relation "\textit{part of}" and \textit{x} a given entity, we can formally describe such a property as shown in \ref{eq:reflexivity}:

\begin{equation} \label{eq:reflexivity}
    P\emph{xx}
\end{equation}

\textit{Transitivity} states a transmission of properties among parts of parts, that is if an entity \emph{x} is a part of an entity \textit{y}, which is in turn part of \textit{z}, then \textit{x} is also a part of \textit{z}, which act as a superordinate whole. This is formalised in \ref{eq:trans}:

\begin{equation} \label{eq:trans}
    P\emph{xy} \land P\emph{yz} \Rightarrow P\emph{xz}
\end{equation}

\noindent{}Therefore, if \ref{ex:transa} and \ref{ex:transb} are true, it follows that \ref{ex:transc} is also necessarily true:

\ex. \label{ex:trans}
\a.\label{ex:transa} \emph{The feather is a part of the wing}
\b.\label{ex:transb} \emph{The wing is a part of the bird}
\b.\label{ex:transc}\emph{The feather is a part of the bird}

\noindent{}While the \textit{reflexivity} property seems to be the less informative about the nature of the relation and its conceptual representation, the \textit{transitivity}, though important, is questioned and poses some theoretical problems that are addressed differently among researchers, who have proposed various solutions to cope with its theoretical treatment. For example, \citet{cruse-lexsem, cruse-transit} distinguishes parts which are \textit{integral} and those which are \textit{attachements} assigning different functional domains to each of them. 
\textit{Attachements} are linguistically distinguished from \textit{integral parts} because the former can fit both in phrasing like \textit{X is a part of Y} and \textit{X is attached to y} as shown in \ref{ex:attach}, while the latter can only sound normal into the \textit{part of} frame, as in \ref{ex:integral}:

\ex. \label{ex:attach-integral}
\a. \label{ex:attach} \textit{The fingers are part of the hand/ The fingers are attached to the hand} (attachment)
\b. \label{ex:integral} \textit{The palm is a part of the hand/ \textbf{*} The palm is attached to the hand} (integral)

Under this assumption, \textit{transitivity} shall not carry over an \textit{attachment} boundary. So, if an entity \textit{X} has parts and the same entity \textit{X} is an \textit{attachment} to an entity \textit{Y}, the parts of \textit{X} are not to be also considered parts of \textit{Y}. For example, while the \textit{fingers are part of the hand}, the \textit{hand is attached to the arm}, and so the \textit{fingers} cannot be said to be part of the \textit{arm} because an \textit{attachment} boundary occurs between the two.

Other authors tackle this issue by drawing a detailed taxonomy of meronymy \citep{Winston1987-taxo, chaffin-emp-taxon}. Under these accounts, several sub-categories of the \textit{part-whole} relation can be listed, as found in \citet{Winston1987-taxo}: 

\begin{enumerate} \label{pw-taxon}
    \item \textit{Component/Integral Object}: handle-cup
    \item \textit{Member/Collection}: tree-forest
    \item \textit{Portion/Mass}: slice-pie
    \item \textit{Stuff/Object}: gin-martini
    \item \textit{Feature/Activity}: paying-shopping
    \item \textit{Place/Area}: oasis-desert
\end{enumerate}

\noindent{}Given such classification, a transitive inference can be valid only when occurring between parts forming a homogeneous hierarchical chain.
Consider for instance the following example:

\ex. \label{ex:wrong-trans}
\a.\label{ex:wrong-transa}\emph{The branch is a part of the tree.}
\b.\label{ex:wrong-transb}\emph{The tree is a part of the forest.}
\b.\label{ex:wrong-transc} \textbf{*}\emph{The branch is a part of the forest.}

In that case, the transitivity along the chain \textit{branch-tree-forest} does not hold, because there are two different meronymic relations at play: One of the kind \textit{Component/Integral object}, holding between \textit{branch} and \textit{tree}, and another one of the kind {Member-Collection} between {tree} and \textit{forest}. Conversely, a single type of meronymic relation, that is \textit{Component/Integral object}, occurs along the chain \textit{feather-wing-bird}, thereby granting the correct application of \textit{transitivity} in \ref{ex:trans}.

The third property of meronymy is \textit{antisymmetry}, which is more intuitive and easier to account for from a theoretical point of view, holding true for every kind of part-whole relation. Antisymmetry claims that \emph{if x is a part of y, then y can not a part of x}:

\begin{equation} \label{eq:meronymy}
    P\emph{xy} \Rightarrow \neg P\emph{yx}
\end{equation}

\noindent{}This property of meronymy explains minimal sentence pairs like those in \ref{ex:minpair}, in which only the former is semantically acceptable, while the latter is anomalous because it violates the antisymmetric constraints: 

\ex. \label{ex:minpair}
\a.\label{ex:minpaira} \emph{The wing is a part of the bird.}
\b.\label{ex:minpairb}\textbf{*}\emph{The bird is a part of the wing.}

The \textit{antisymmetry} property seems to be immune from the problems posed by \emph{transitivity} and always hold true, regardless of the specific part-whole relation taken into account. Therefore, in this work we decided to focus only on the \emph{antisymmetry} of meronymy, leaving aside \emph{reflexivity} for its low degree of informativeness and \emph{transitivity} for its peculiar and unclear theoretical status.
Though the antisymmetry of meronymy is fairly trivial for humans to grasp, it is important to explore its mastering by computational models relying only on statistical correlations among words to build their representations of the world, which may be highly biased and unstable \citep{kang-choi-2023-impact}.

\section{Experiments overview and relevant works} \label{sec:exp-overview}

Despite its theoretical relevance, the \emph{part-whole} relation is significantly understudied in NLP, let alone in the general quest for understanding the semantic capabilities of LLMs. To fill this gap, we conducted a three-pronged investigation revolving around the following methodological pillars: i) \textbf{behavioral analysis} with prompting, ii) \textbf{probabilistic  analysis} of sentence plausibility, and iii) \textbf{representational analysis} of the part-whole relation in the embedding space. We regard these three levels of inquiry as complementary and synergistically contributing to get at better understanding of the ``knowledge'' of the \textit{part-whole} relation in LLMs.
Broadly speaking, this work aims at contributing to the ongoing debate on how LLMs represent and process factual information and ontological relations between objects of the actual world. The \textit{part-whole} is indeed a crucial relation to understanding real-world configurations and to properly reason about objects \citep{Gerstl_Pribbenow_1995}.
In the remainder of this section, we outline our methods and briefly discuss some of the relevant literature.

\textbf{Behavioral analysis.} 
Comparing the behavioral responses by artificial systems to human ones has long been a privileged tool for making assumptions about their internal knowledge and capabilities \citep{Turing1950-TURCMA}. Indeed, authors have argued that this is the right way to test AI models and draw conclusions about their level of intelligence \citep{Levesque2009IsIE, Levesque2014OnOB}.
With the advent of LLMs, \textbf{prompting} has lately become a widely used technique for their behavioral analysis and to make them accomplish several tasks \citep{brown2020}. This trend has become even more popular with models fine-tuned to follow instructions \citep{Ouyang-instruction-follow, chung2022scalinginstructionfinetunedlanguagemodels}. This has contributed to spark interest in treating LLMs as kinds of psycholinguistics subjects \citep{futrell-etal-2019-neural}. In fact, prompting is an appealing tool to operate linguistic analysis in a fast and easy way \citep{li-etal-2022-probing-via, blevins-etal-2023-prompting}, opening up the possibility to interact with LLMs like with human subjects in standard (psycho)linguistic experiments, without the burden of complicated technical settings. Indeed, prompting allows researchers to transfer directly to LLMs the current psycholinguistics experimental toolbox, provided adequate adaptation. Several works which have incorporated prompting in their analysis are relevant to ours.
For instance, \citet{Hansen2022} used GPT-3 to generate semantic features of objects following \citet{McRae:2005} for items contained in the Things dataset \citep{Hebart2019} (see below, Section \ref{sec:data}).
\citet{berglund2024reversal-curse} showed that LLMs may fail at learning symmetric relational properties, thus lacking generalization capabilities, a problem they dubbed the \textit{reversal curse}. This is relevant to our work because we also explore the ability of LLMs to capture the semantic properties of meronymy. However, they focus on the identity relation (\emph{Neil Armstrong was the first man on the moon} = \emph{The first man on the moon was Neil Armstrong}), which is inherently symmetric, while meronymy is an antisymmetric relation.
On the same track, \citet{qi-etal-2023-investigation} have conducted experiments to assess the abilities of LLMs to correctly make inferences about \textit{converse relations} focusing on $17$ relations. The goal of both these works is to gain evidence about the ability of LLMs to understand semantic relations and generalize correct inference patterns about them. There are few works dealing directly and exclusively with the processing of the \emph{part-whole} relation in LLMs. \citet{gu-etal-2023-lms-mental-mod} investigated whether LLMs possess a coherent model of common objects by asking them to classify what relation occurs between parts of a given object. Although related to ours, their work more specifically focuses on \emph{co-meronymy}, which deals with relations among parts of a given whole and their configuration, rather than strict meronymy, which is focused on the relation between parts and their wholes.

\textbf{Probabilistic analysis.} Behavioral analysis via prompting has its own intrinsic limitations. While it can be taken as a demonstration of the abilities the models have in generating certain outputs, it falls short of being conclusive about the kind of information which may be encoded in LLMs. For example, \citet{hu-levy-2023-prompting} stresses the discrepancy between results obtained through asking metalinguistic judgments via prompting and the inspection of the model probability space. Following that and similar recommendations, we tried to counterbalance the shaky explanatory power of prompting analysis by taking into account measurements in models' probability space to evaluate \textbf{sentence semantic plausibility}. Taking probability scores assigned to sentences by the models is another popular way to evaluate their ability to discriminate correct and incorrect sequences along the dimensions of grammaticality \citep{marvin-linzen-2018-targeted} and semantic plausibility \citep{kauf_etal:2023,kauf-etal-2024-log}. This is usually obtained by constructing minimal pairs of acceptable and unacceptable sentences and comparing the scores assigned to them by the model \citep{warstadt-etal-2020-blimp-benchmark, gauthier-etal-2020-syntaxgym}. \citet{wiland-etal-2024-bear} recently presented a unified framework to evaluate relational knowledge in language models, on the track initiated by \citet{petroni-etal-2019-language} with probability estimation and alternation of minimally different inputs.
In our specific case, evaluation examples are constructed by pairing a correct sentence stating a part-whole relation (\textit{The wheel is part of the car}) with a counterfactual represented by the swapped version of the former (\textit{The car is a part of the wheel}). To the best of our knowledge, this technique has not been directly employed yet to evaluate models' on their ability to handle \emph{part-whole} relations.

\textbf{Representational analysis.} Since the seminal work by \citet{mikolov-etal-2013-linguistic}, it has been widely claimed that concepts are encoded linearly in the embedding spaces of distributional semantic models \cite{Lenci2023}. Recently, this \textbf{linear representation hypothesis} (LRH) has also be revived for LLMs \citep{park2024linearrepresentationhypothesisgeometry, Jiang2024OnTO, Park2024TheGO}. According to the LRH, concepts form linearly separable sub-spaces inside models' vector and activation spaces. If true, this hypothesis could serve as a viable tool to make assumptions on the conceptual and lexical organization in linguistically derived vector spaces. Following the LRH, we extracted static embeddings from the input and output layers of the models, searching for some linearity in the structural organization of a possible \textit{part-whole} concept in the vector spaces (embedding and unembedding) of the models, applying methods laid out in \citet{park2024linearrepresentationhypothesisgeometry}. Interestingly, among the $27$ concepts analysed in \citet{park2024linearrepresentationhypothesisgeometry} the \textit{part-whole} is the only one declared to be non-linearly encoded by the authors. We used a set of newly created benchmarks (cf. Section \ref{sec:data}) to further test their claims concerning this particular relation and the suitability of the LRH to analyse its encoding by LLMs.\newline



\section{Experimental items} \label{sec:data}
\begin{table}[t]
{%
\begin{tabular}{lccc}
\textbf{Dataset} & \textbf{Holonyms} & \textbf{Meronyms} & \textbf{Lemmatized} \\ \hline
\textsc{mcrae} & 424 & 998 & No \\
\textsc{mcrae\_lemma} & 424 & 994 & Yes \\
\textsc{conceptnet} & 576 & 1,026 & Yes\\
\hline
\end{tabular}%
}
\caption{Experimental items and datasets.}
\label{tab:datasets}
\end{table}

In our experiments, we used three datasets (see Table \ref{tab:datasets}) consisting of \textsc{<meronym, holonym>} pairs collected from the following sources:

\begin{itemize}
    \item \textbf{Mc Rae's Feature Norms}:
    \citet{McRae:2005} elicited semantic feature norms for a set of 541 basic concepts, comprising living and non-living entities, from a group of 725 participants, asked to tell semantic properties of the target objects.
    
    \item \textbf{ConceptNet}: this is a large multilingual knowledge graph consisting of words and short phrases connected by commonsense relations derived from various sources (e.g., such as   Wiktionary, the Open Mind Common Sense project, games with a purpose, etc.; \citealt{speer2016conceptnet}).

    \item \textbf{Things}:
    this dataset contains a list of 1,854 objects concepts, both living and non-living, associated with more than 20k images describing them \citep{Hebart2019}.
\end{itemize}

From the McRae's norms, we extracted each human judgment labeled as \texttt{external\_component} and marked with the \texttt{has\_<part>}, we removed digits or adjectives where present, and reversed the nodes (e.g., <\texttt{dog}, \texttt{has\_a\_long\_tail}> $\rightarrow$ <\texttt{tail,dog}>). After this normalization step, we got a list of <meronym,holonym> pairs from which  we built two separate datasets: \textsc{mcrae}, with the original meronyms in \citep{McRae:2005}, and \textsc{mcrae\_lemma} with a lemmatized version of the same data.\footnote{The data were lemmatized with the \texttt{spaCy} python library \citep{honnibal2020spacy}, using the large English pipeline (\texttt{en\_core\_web\_lg}), and then manually revised to correct faulty outputs.}\break
The third dataset, \textsc{conceptnet}, was obtained by gathering holonyms from the union of the objects list in the McRae's norms and the Things dataset, whose meronyms were then obtained by querying the ConceptNet graph for the \texttt{partOf} relation.\footnote{For consistency with the linguistic parsing tool (SpaCy) we used \texttt{concepcy}, a ConceptNet wrapper for the \texttt{SpaCy} python library}

\section{Models}\label{sec:models}
Nowadays, the number of extant LLMs is growing exponentially and the community behind their development is constantly releasing new and updated versions of them. This proliferation is made possible and easy by the \textit{pre-train and fine-tune} paradigm. In this setting, the refinement of a model through \textit{fine-tuning}  could virtually be iterated \textit{ad libitum}, causing the spring of a multitude of models from both a common pre-trained base and/or already fine-tuned versions of it.
However, in spite of the increasing swarm of LLMs available to the public, models often differ only slightly from each other with respect to the architecture and training paradigm.
Given these assumptions, in this work, we did not intend to conduct a thorough comparison of LLMs, competitively benchmarking their abilities. Instead, we opted to select a small representative of the major types of autoregressive decoder-only LLMs available on the scene today. Moreover, our choice is also dictated by the computing resources at our disposal at the time of the experiments.
Specifically, we tested the following three LLMs:
\begin{itemize}
    \item \textbf{LlaMA2-7b}.\footnote{https://huggingface.co/meta-llama/Llama-2-7b} An open-source, autoregressive decoder-only pre-trained language model released by META \citep{touvron2023llama}. It is in the range of 7 billion parameters and is trained on 2 trillion tokens with a context window of 4 thousand tokens. It is composed of 32 transformer blocks, each having 32 attention heads in its attention layer.
    \item \textbf{LlaMA2-7b-chat}.\footnote{https://huggingface.co/meta-llama/Llama-2-7b-chat-hf} This is the chat-specialized version of the preceding model, with the same architectural characteristics \citep{touvron2023llama}. This model has undergone a supervised fine-tuning phase and it has been aligned to human preferences through reinforcement learning with human feedback \citep{chris-rlhf}.
    \item \textbf{GPT-4}. The latest stable version (non-preview) of the GPT family \citep{openai2024gpt4}. Differently form the previous ones, this is a proprietary model and, unfortunately, not much architectural information is available. 
\end{itemize}

\begin{table}[t]
\begin{tabular}{l|cc|c|c}
\multirow{2}{*}{\textbf{Model}} & \multicolumn{2}{c|}{\textbf{\begin{tabular}[c]{@{}c@{}}Behavioral \\ Analysis\end{tabular}}} & \multirow{2}{*}{\textbf{\begin{tabular}[c]{@{}c@{}}Probabilistic \\ Analysis\end{tabular}}} & \multirow{2}{*}{\textbf{\begin{tabular}[c]{@{}c@{}} Representational\\
Analysis
\end{tabular}}} \\ 
 & \multicolumn{1
 }{c|}{Task1} & Task2 &  &  \\ \hline
LlaMa2-7b & \multicolumn{1}{c|}{\checkmark} &  & \checkmark & \checkmark \\ 
LlaMa2-7b-chat & \multicolumn{1}{c|}{\checkmark} & \checkmark & \checkmark & \checkmark \\ 
GPT-4 & \multicolumn{1}{c|}{\checkmark} & \checkmark &  &  \\
\hline
\end{tabular}%

\caption{Task distribution per model. Different model are tested on different tasks, depending on their being open- vs. closed-source and instruct vs. non-instruct.}
\label{tab:tasksxmodel}
\end{table}

Due to their differences in architecture, training regimes and openness, not all the models are suitable for the tasks we devised.
Table \ref{tab:tasksxmodel} gives a detailed overview of which model was tested on which task.

\section{Behavioral Analysis} \label{behav-task-desc}

We performed the behavioral analysis of meronymy in LLMs with three prompting tasks designed to probe their knowledge of the \emph{part-whole} relation and its logico-semantic properties:\newline

\noindent{}\textbf{Task 1: meronymy understanding} -- This task consists in directly querying the LLMs whether the items  in the three datasets described in Section \ref{sec:data} are instances of meronymy. The task comes in two versions:
\begin{enumerate}
    \item \textbf{Binary question answering}: the model is asked to answer binary YES/NO, questions, such as \emph{Is the wheel a part of the car?}
    \item \textbf{Binary statement verification}: the model is asked to assign a truth value to a meronymic statement, such as \emph{The wheel is a part of the car.}
\end{enumerate}
In order to test whether models know that meronymy is an antisymmetric relation, the sentences correctly answered by the LLMs were fed them back to them in a \textbf{swapped counterfactual version} (i.e., \emph{Is the car a part of the wheel?/The car is a part of the wheel}). We performed the swapped test only on the correct answers, to be sure that the model has some knowledge of the meronymy relation between two entities, and only then check whether it treats such relation as antisymmetric. The original and swapped tests were used to define the following \textbf{Meronymy Knowledge Criterion}:

\ex. \label{MKC} A LLM knows that a pair $\langle$x,y$\rangle$ is an instance of meronymy iff the LLM correctly solves the direct and the swapped test for the pair $\langle$x,y$\rangle$.

\begin{figure}
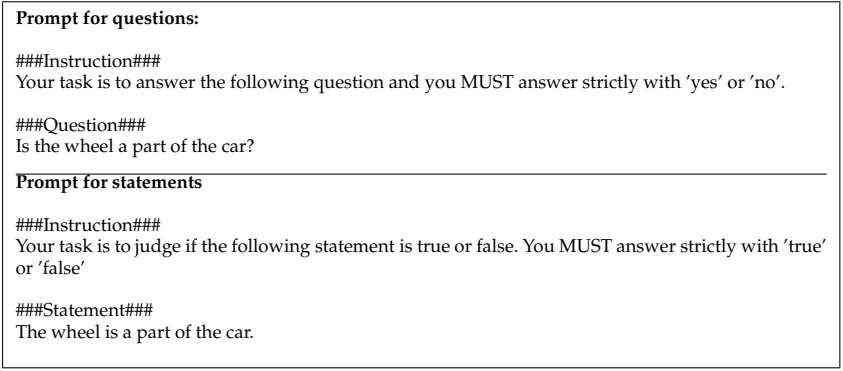

\centering
\scriptsize
\fbox{
	\parbox{0.8\linewidth}{
        \textbf{Prompt for questions:}\\
        \\
	\#\#\#Instruction\#\#\# \\
        Your task is to answer the following question and you MUST answer strictly with 'yes' or 'no'.\\\\ 
        \#\#\#Question\#\#\#\\
        Is the wheel a part of the car?\\
        \hrule 
        \textbf{Prompt for statements}\\
        \\
	\#\#\#Instruction\#\#\# \\
        Your task is to judge if the following statement is true or false. You MUST answer strictly with 'true' or 'false'\\\\
        \#\#\#Statement\#\#\#\\
        The wheel is a part of the car.\\
	}
 }
 \caption{Prompt templates used for Task 1: meronymy understanding.}
 \label{prompt1:0-shot}
\end{figure}

\noindent{}Both versions of Task 1 were performed in a \emph{0-shot} setting for the instruct models, LlaMA2-7b-chat and GPT4, and in a \emph{5-shot} setting for the non-instruct model, LlaMA2-7b. The prompts were minimally different between LlaMA2-7b-chat and GPT-4. While for GPT-4 they were exactly as shown in Figure \ref{prompt1:0-shot}, for LlaMA2-7b-chat the system-prompt (i.e., the instruction) was wrapped within the \texttt{<<SYS>>} and \texttt{<</SYS>>} tokens, to signal the beginning and the end of the system prompt, and the whole text was wrapped within \texttt{[INST]} and \texttt{[/INST]} to tell the model the end of the complete instruction, as suggested by META researchers and best practices. For the non-instruct model, we replaced the instruction with 5 examples as shown in Figure \ref{5-shot prompt}. To make the prompts as effective as possible, we incorporated two principles from \citet{bsharat2024principled}, specifically the 8 and 9 principles, concerning the clearness and conciseness of the prompt formulation.\newline

\begin{figure}
\centering
\scriptsize
\fbox{
	\parbox{0.8\linewidth}{
        \textbf{5-shot Prompt for questions:}\\
        \\
	Question: Is the wing a part of the aeroplane?\\
        Answer:yes.\\
        Question: Is the aeroplane a part of the wing?\\
        Answer: no.\\
        Question: Are the legs a part of the car?\\
        Answer: no.\\
        Question: Are the seeds a part of the zucchini?\\
        Answer: yes.\\
        Question: Is the sword a part of the blade?\\
        Answer: no.\\
        Question: Is the tail a part of the dog?\\
        Answer:\\
        \hrule 
        \textbf{5-shot Prompt for statements}\\
        \\
        Statement: Is the wing a part of the aeroplane?
        \\Judgement: TRUE.\\        
        Statement: Is the aeroplane a part of the wing?
        \\Judgement: FALSE.\\        
        Statement: Are the legs a part of the car?
        \\Judgement: FALSE.\\        
        Statement: Are the seeds a part of the zucchini?\\Judgement: TRUE.\\        
        Statement: Is the sword a part of the blade?
        \\Judgement: FALSE.\\        
        Statement: The tail is a part of the dog.\\
        Judgement:\\
	}
 }
 \caption{Prompt template in 5-shot used for LlaMA2-7b in the first task}
 \label{5-shot prompt}
\end{figure}



\noindent{}\textbf{Task 2: part generation} -- 
The goal of this task is to test the models' ability to generate parts of target concepts. We fed the LLMs with an object name along with an instruction to list its parts (see Figure \ref{prompt:partgen}). This task was performed only with the instruct models. Like in Task 1, the only difference in the prompt is the presence of the special tokens required by LlaMA2-7b-chat.\newline

\noindent{}\textbf{Task 3: self-generated meronymy understanding} -- This is the exact replication of Task 1, and the only difference is that now the test items are \textsc{<meronym, holonym>} pairs containing the parts generated by the LLMs themselves (after being manually cleaned).
\break{}

\begin{figure}

\centering
\scriptsize
\fbox{
	\parbox{0.8\linewidth}{
        \textbf{Prompt for generations:}\\
        \\
	\#\#\#Instruction\#\#\# \\
        Your task is to list the parts of a given object. You MUST format the output as a comma-separated list.\\\\ 
        \#\#\#Object\#\#\#\\
        Car.
        \\
	}
 }
 \caption{Prompt template for Task 2: part generation.}
 \label{prompt:partgen}
\end{figure}

\noindent{}In all tasks, the LLMs have been used in inference mode, without any fine-tuning phase with the temperature parameter set to zero.

\subsection{Results for Task 1: meronymy understanding} \label{task1-res}


 %
\begin{figure}
    \centering
    \includegraphics[scale = 0.5]{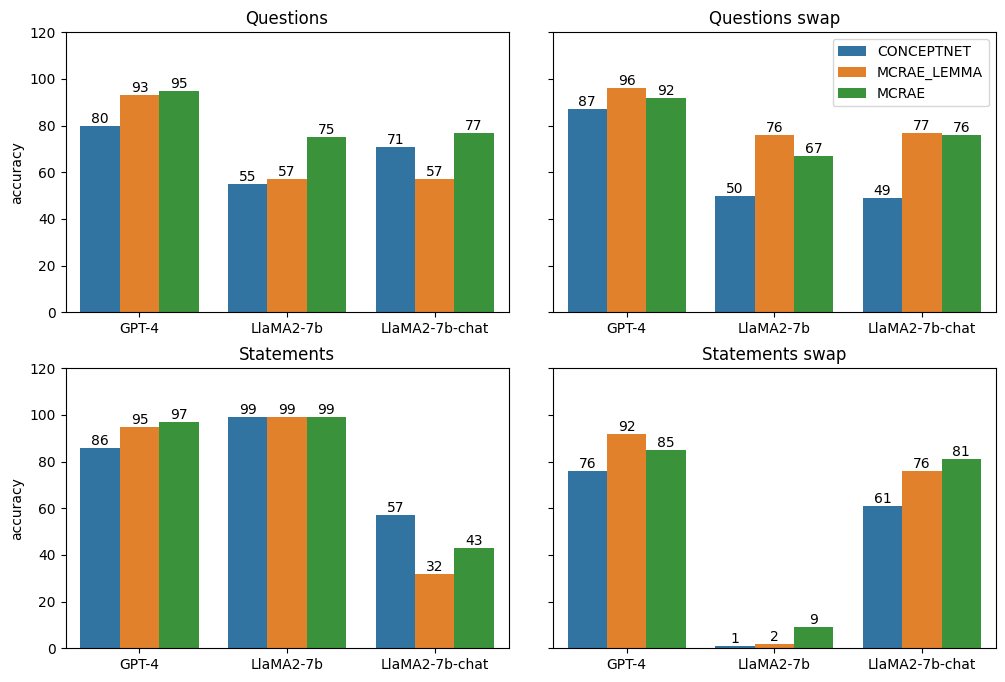}
    \caption{Models' accuracy for Task 1: question answering (top line), statement verification (bottom line), original pairs (left column), swapped items (right column)}
    \label{fig:task1-total}
\end{figure}

\begin{figure}
    \centering
    \includegraphics[scale = 0.5]{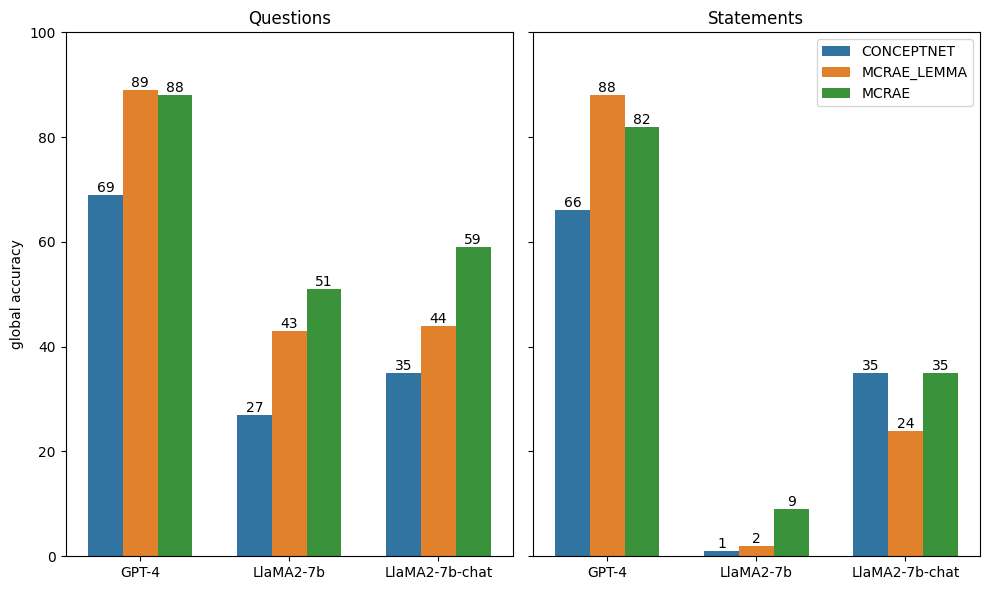}
    \caption{Global accuracy of the LLMs in satisfying the Meronymy Knowledge Criterion.}
    \label{fig:task1-global}
\end{figure}

Figure \ref{fig:task1-total} shows the accuracy of the LLMs in recognizing the test pairs in the three datasets as instances of meronymy. It is worth remarking that in the swapped tests the accuracy is computed with respect to the correct answers produced by the models in the direct version (e.g., an accuracy of $75\%$ in the swapped test means that only three quarters of the correctly answered direct prompts were also answered correctly in the swapped version). Figure \ref{fig:task1-global} reports the accuracy of the LLMs in satisfying the Meronymy Knowledge Criterion in \ref{MKC}.


As can been seen in Figure \ref{fig:task1-total}, LLMs are very good in solving the direct meronymy understanding tasks (left column), except for LlaMA2-7b-chat in the statement verification version. However, if we consider the Meronymy Knowledge Criterion in Figure \ref{fig:task1-global}, the models' accuracy drop significantly. LlaMA-7b-chat and its non-instruct counterpart, LlaMA-7b, show poor performances around or much below chance level. Unsurprisingly, GPT-4 scores much better than the other models, but on the \textsc{conceptnet} dataset its performance is just $69\%$ on the question answering task and $66\%$ on the statement verification one. 
Generally, models seem to struggle more with the ConceptNet data. The meronyms in ConceptNet are more specific, technical and uncommon than those elicited from people in the feature norming task conducted by \citet{McRae:2005}. This difference might explain the consistent gap in the performance of the LLMs between the ConceptNet and McRae data.

\subsection{Results for Task 2: part generation} \label{gen-res}
\begin{figure}[t]
    \centering
    \includegraphics[scale = 0.7]{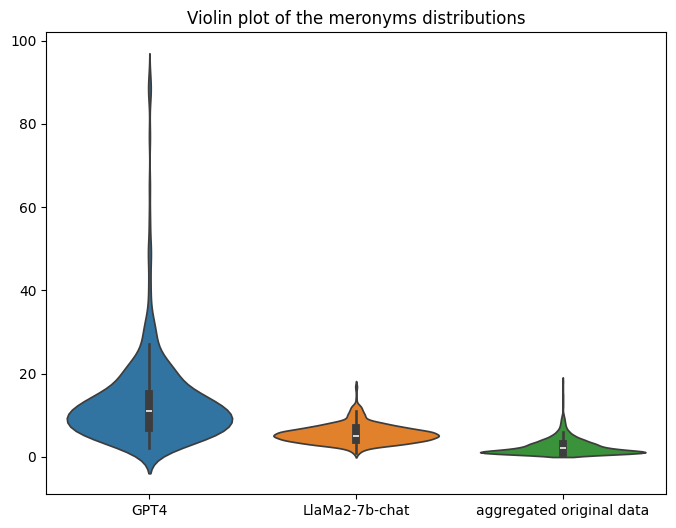}
    \caption{Distribution of meronyms for each holonym as generated by each model, together with thri distribution in the aggregated \textsc{conceptnet} and \textsc{mcrae} datasets.}
    \label{fig:violin}
\end{figure}

This task was performed only with the two instruct models. In general, both LLMs followed the instructions and generated several parts of the holonyms in the prompts (see Table \ref{tab:partgen}). However, for LlaMA2-7b-chat a thorough cleaning of the output was needed to remove noisy, wrong and unnecessary generations. 
After this process, we had 4,242 correct meronyms over 761 holonyms, with an average of 6 generated parts for each input object.
The parts generated by GPT-4 were much richer and cleaner, for a total of 11.5k items. After a shallow formal cleaning, we got rid of some instances, reducing the whole generated outputs to 11,627 parts for 847 objects, with an average of 14 parts generated per object, which is more than two times those of LlaMA2. As can be seen in Figure \ref{fig:violin}, the distribution of meronyms is pretty skewed on the lowest range for each dataset. However, the GPT4-generated parts show a wider distribution per holonym with a tail of few objects with an increasing number of generated parts.

The part generation ability of the LLMS is extremely high. All the parts generated by LlaMA2-7b-chat were manually checked at the semantic level, reporting an accuracy of $92.9\%$. For GPT-4, we manually checked 300 randomly selected holonyms, for a total of 4k meronyms, obtaining an accuracy of $97.5\%$.
However, we also got errors, which were different across models, 
with LlaMA2-7b-chat showing a greater variety of mistakes. We will further discuss the main types of errors in Section \ref{sec:wrap-up}


\begin{table}[t]
\begin{tabular}{lcccc}
\textbf{Model/Data} & \textbf{Total} & \textbf{Min} & \textbf{Avg} & \textbf{Max} \\ \hline
\textsc{LlaMA2-7b-chat} & 4,242 & 1 & 6 & 17 \\
\textsc{GPT-4} & 11,627 & 2 & 14 & 91 \\ \hline
\textsc{conceptnet+mcrae} & 1,999 & 1 & 2 & 18\\
\hline
\end{tabular}%
\caption{Summary statistics on the part generation task, in comparison with the aggregation of the original datasets, with minimum, maximum, and average generated parts per holonym.}
\label{tab:partgen}
\end{table}

\subsection{Results for Task 3: self-generated meronymy understanding } \label{self-gen-res}

Given the output of Task 2, we created two additional datasets composed of 4,242 and 11,627 \textsc{<meronym,holonym>}, generated respectively by LlaMA2-7b-chat and GPT4. Then, we replicated the methodology in Task 1 on these datasets, to test whether the models were able to recognize as instances of meronymy their own generations.

\begin{figure}[t]
    \centering
    \includegraphics[scale = 0.5]{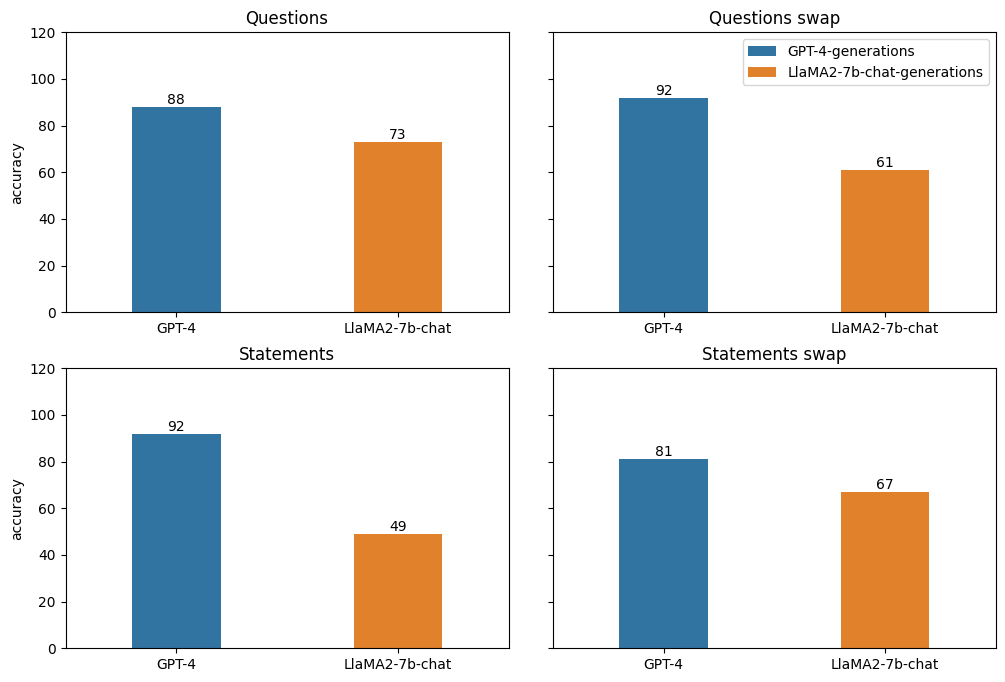}
    \caption{Models' accuracy for Task 3 on self-generated parts: question answering (top line), statement verification (bottom line), original pairs (left column), swapped items (right column).}
    \label{fig:res from gen tot}
\end{figure}

\begin{figure}[t]
    \centering
    \includegraphics[scale = 0.5]{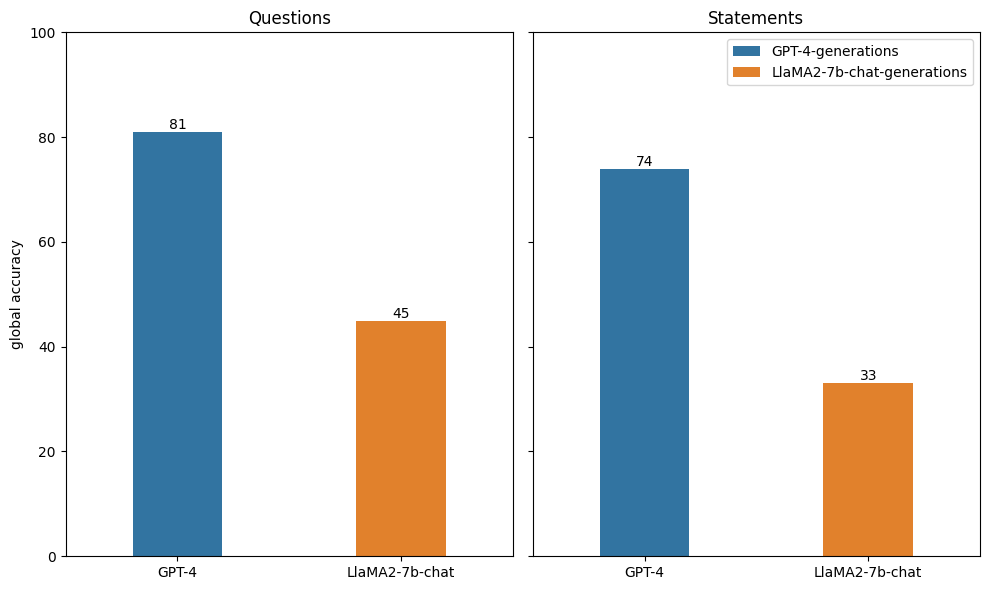}
    \caption{Global accuracy of the LLMs in satisfying the Meronymy Knowledge Criterion on self-generated parts.}
    \label{fig: res_from_gen}
\end{figure}

As shown in Figure \ref{fig:res from gen tot}, the performances of the models on their own generated parts seem to partially follow the trend reported for Task 1, with GPT-4 again showing much higher performances than LlaMA2-7b-chat. Interestingly, although models were dealing with data they generated in the first place, there was no significant improvement over the best results obtained in Task 1. Figure \ref{fig: res_from_gen} strikingly shows that the Meronymy Knowledge Criterion was not satisfied by LlaMA2-7b-chat for the large majority of its self-generated parts, and that even the top-performing GPT-4 was able to match this criterion only for $81\%$ of its own parts in the question answering task, a score that goes down to $75\%$ in the statement verification task.

\section{Probabilistic Analysis} \label{sec:sent-plaus}
The probability scores assigned by the two open-source language models (LlaMa2-7b and LlaMa2-7b-chat) to meronymic statements were used to estimate their knowledge of the \emph{part-whole} relation. From the meronymic pairs in the three datasets employed in the behavioral analysis, \textsc{conceptnet}, \textsc{mcrae} and \textsc{mcrae\_lemma} meronyms, we defined a set of pairs $\langle m, m^{-1}\rangle$, where $m$ is a meronymic statement (i.e., \textit{The wheel is a part of the car}) and $m^{-1}$ is its swapped version (i.e., \textit{The car is a part of the wheel}). We fed the models with these statements and we collected their log probability scores. As shown by \citet{kauf_etal:2023}, log probability is a good estimator of a sentence semantic plausibility and we reformulated the \textbf{Meronymy Knowledge Criterion} in \ref{MKC} as follows:

\ex. \label{MKC_1} A LLM knows that a pair $\langle$x,y$\rangle$ is an instance of meronymy iff, given the corresponding sentence pair $\langle m, m^{-1}\rangle$, the LLM assigns to $m$ a greater log probability than to $m^{-1}$ (i.e., $log P(m) > log P(m^{-1}$).



We measured the model's accuracy in recognizing meronymy as the percentage of times it assigns a greater log probability to $m$ than to $m^{-1}$. As a baseline, we built a set of input pairs in which a fake \emph{part-whole} relation is stated between random objects (e.g., \textit{The zebra is a part of the table}/\textit{The table is a part of the zebra}).  In this case, we expected the models to perform around the chance level.

\begin{figure}
    \centering
    \includegraphics[width= \textwidth]{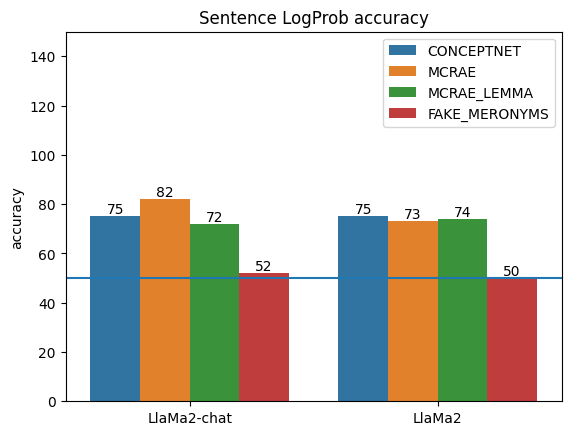}
    \caption{Accuracy of the models assigning log probabilities to the sentences}
    \label{fig:sents-prob}
\end{figure}

As shown in Figure \ref{fig:sents-prob}, both LLMs perform significantly over the baseline with an average percentage of correct answers around $75\%$. In general, accuracy is better than in the behavioral, prompt-based tasks, and there is no significant difference between non-instruct and instruct models. However, it is striking to observe that almost one quarter of semantically plausible meronymic statements are not regarded as being more likely than their implausible swapped versions.

\section{Representational Analysis}\label{linhypo-method}


We analysed the \textit{embedding} and \textit{unembedding} layers of the LLMs, in order to understand how the \textit{part-whole} relation is encoded in the input and output representations of the Transformer architecture \citep{vaswani-transformer}, which is at the heart of such models.

Two particular layers are usually posed at the beginning and at the end of the whole LLM, governing the input (encoding) and output (decoding) representations respectively: the matrix $\textbf{W}_E \in \mathbb{R}^{|\mathcal{V}|\times d}$, called the \textbf{embedding} layer, and the matrix $\textbf{W}_U \in \mathbb{R}^{d\times |\mathcal{V}|}$, called the \textbf{unembedding} layer, where $\mathcal{V}$ is the set of tokens forming the model's vocabulary and $d$ is the dimension of the inner representation. 
The tokens of an input sentence are encoded into their embedding representations, then pass across the network layers getting updated by the attention mechanism and the other components of the Transformer blocks. The representation updated through the network is called the \textbf{residual stream} \citep{elhage2021mathematical, Ferrando2024APO}, and once it reaches the last layer it is decoded into the vocabulary space by the unembedding matrix $\textbf{W}_U$. This step transforms the residual stream into logits that the softmax function then maps onto a probability distribution of the next token to be predicted.

Both embedding layers contain a representation of the models' vocabulary acquired during the pre-training phase. However, the semantic information actually encoded in such representations is notoriously opaque, and the way it is organized across the whole space is unclear. In order to understand how the \emph{part-whole} relation is represented in the models' embedding space, we extended the analysis in \citet{park2024linearrepresentationhypothesisgeometry}, whose theoretical and methodological foundations are grounded in the so-called \textbf{linear representation hypothesis}. This assumption first advanced by \citet{mikolov-etal-2013-linguistic} claims that the embedding spaces acquired by distributional semantic models are organized in terms of liner subspaces corresponding to high-level concepts. 

\citet{park2024linearrepresentationhypothesisgeometry} define a \textbf{concept} as a semantic dimension that discriminates one word from another and formalize it with word pairs whose components are distinguished by that dimension. For example. the \textit{gender} concept is characterized with pairs like $\langle man, woman \rangle$ and $\langle king, queen \rangle$. Analogously, the \textit{part-whole} relation defines the \textit{partOf} concept, which is expressed by word pairs such as $\langle wheel, car \rangle$ and $\langle wing, aircraft \rangle$. According to the linear representation hypothesis, word pairs expressing the same concept share similar vector offsets \citep{mikolov-etal-2013-linguistic}. For instance, the difference between the \textit{man} vector and the \emph{woman} vector is expected to parallel the difference between the \emph{king} and \emph{queen} vectors. Therefore, this hypothesis assumes that concepts are represented in the embedding spaces as \textbf{directions} defined by the word pairs that express them.

In order to gain insights about the semantic structure in the LLMs' vector spaces, \citet{park2024linearrepresentationhypothesisgeometry} proposed a method based on the linear representation hypothesis that they tested against $27$ concepts, spanning different dimensions and derived from the BATS benchmark \citep{gladkova-etal-2016-analogy}. Although the \textit{partOf} concept is one of those, the word pairs that represented it were only $13$ and corresponded to very different subtypes of the \textit{part-whole} relation (cf. Section \ref{part-whole-rel}). We applied their methodology to our much larger dataset by defining 
the \textit{partOf} concept as the set $Z =\{\langle m_1,h_1\rangle,\dots\,\langle m_n,h_n \rangle\}$, where each $\langle m_i,h_i \rangle$ represents a meronymic pair used in the experiments in Sections \ref{behav-task-desc} and \ref{sec:sent-plaus}. The procedure by \citet{park2024linearrepresentationhypothesisgeometry} was adapted in the following way:
\begin{enumerate}
    \item We extracted the embedding ($\textbf{W}_e$) and unembedding ($\textbf{W}_u$) representations of both LlaMA2-7b and LlaMA2-7b-chat and applied a whitening transformation to remove correlations, thus deriving centered matrices $\mathbf{\Gamma_e}$ and $\mathbf{\Gamma_u}$, such that 
    $\mathbf{\Gamma} = (\mathbf{W} - \bar{\mathbf{w}}) \left[\frac{1}{W}(\mathbf{W} - \bar{\mathbf{w}})^\top (\mathbf{W} - \bar{\mathbf{w}})\right]^{-\frac{1}{2}}$

    \item We selected the embeddings $\gamma$ of 1,742 meronymic pairs from the union of the \textsc{mcrae} and \textsc{conceptnet} datasets, and 3,249 pairs from the LlaMa2-7b-chat self-generations. We aggregated sub-word representations for those items split by the tokenizer,\footnote{An objection to this choice might be that we are not dealing directly with concepts naturally encoded in the vector space. However, if LLMs do encode concepts in linear subspaces, this should not be restricted only to concepts expressed by single-token words. Although the process of aggregating vectors will likely introduce some noise, it has the advantage of guaranteeing a greater pool of test word pairs.} while we ruled out multi-word expressions.
    
    \item The assumption is that the \emph{partOf} concept is represented as a direction in the space and, given the set $Z$ of meronymic pairs, the direction vector $\hat{\gamma}(partOf)$ is computed as the average of the vector differences of target pairs elements: $\hat{\gamma}(partOf) = \frac{1}{nZ}\sum_{i=1}^{nZ}[\gamma (m_i) - \gamma(h_i)]$ where $m_i$ and $h_i$ refer respectively to the meronym and the holonym element of the $i_{th}$ pair. Like in \citet{park2024linearrepresentationhypothesisgeometry}, we computed this in the leave-one-out fashion: For each pair $i$ in our dataset of size \textit{n} we have the concept direction computed on \textit{n}- \textit{i} elements.
    
    \item Given the concept direction $\hat{\gamma}$, we computed the dot product between the vector difference of a given target pair $i$ and the vector representing the \emph{partOf} concept to see if they align and point toward similar directions:
    $\hat{\gamma}(partOf) \cdot (\gamma(m_i) - \gamma(h_i))$.
    
    \item We compared the distribution of the dot products between the target pair differences and the direction vector $\hat{\gamma}(partOf)$ against the dot products between the same concept vector and set of randomly selected pairs.
\end{enumerate}

\noindent{}If the linear representation hypothesis holds true and the \emph{part-of} is represented as a linear (un)embedding subspace, we expect target pairs to be significantly more aligned with the direction vector than the random pairs.

\subsection{Results: representational analysis}

\begin{figure}
    \centering
    \includegraphics[width = \textwidth]{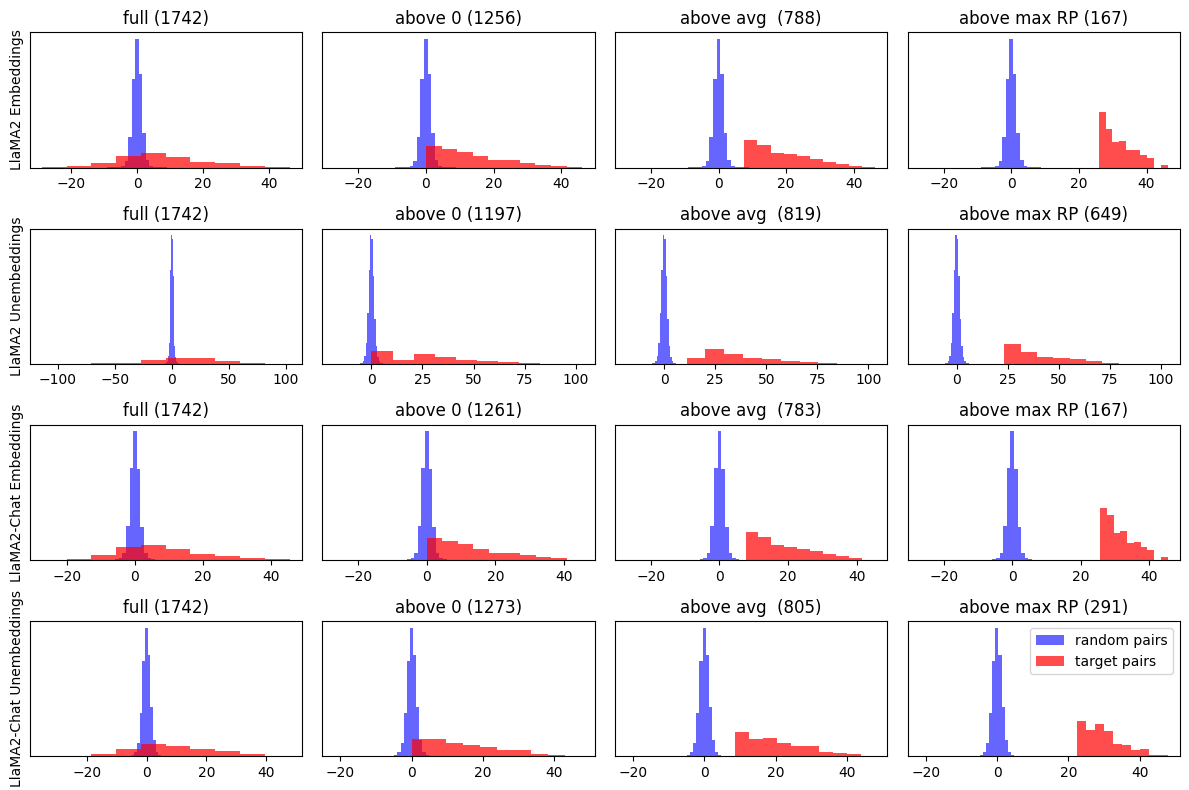}
    \caption{Comparison between the original distribution of products and those obtained with different thresholds for the \textsc{MCRAE} and \textsc{CONCEPTNET} aggregated data}
    \label{fig:thresholds-inputdata}

    \centering
    \includegraphics[width = \textwidth]{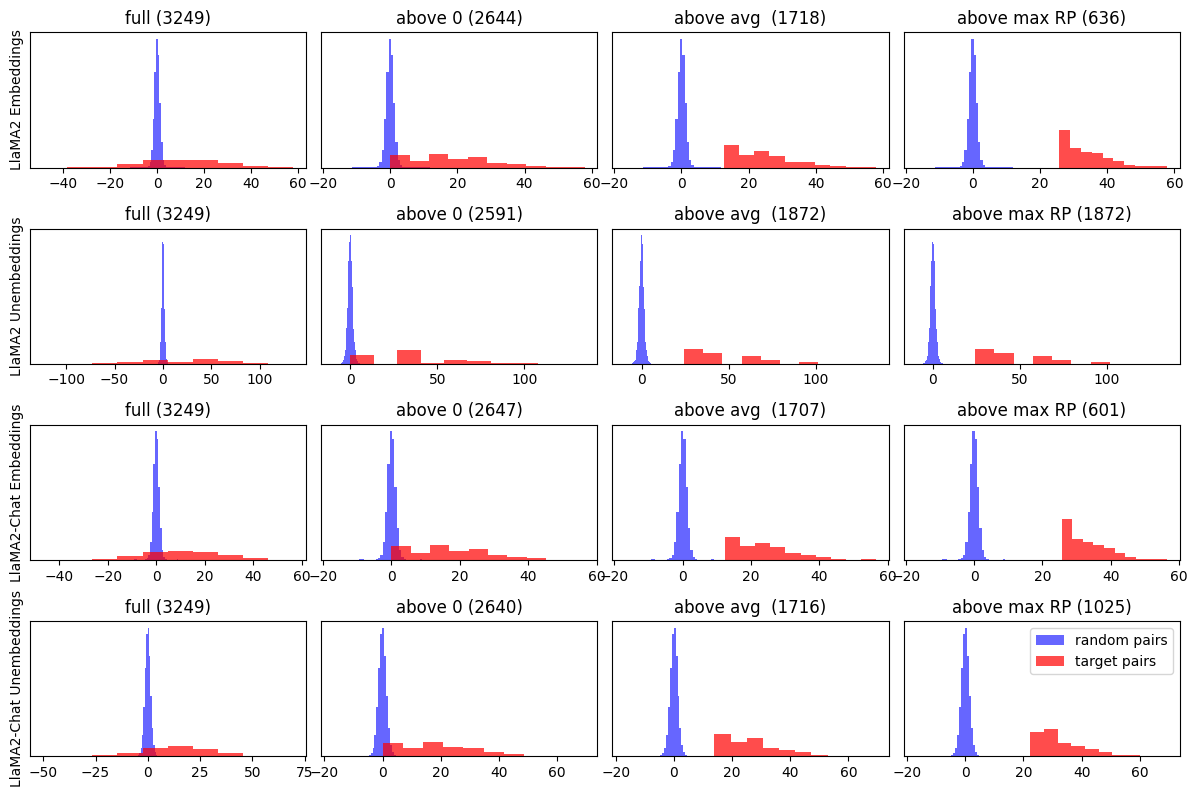}
    \caption{Comparison between the original distribution of products and those obtained with different thresholds for the meronymic pairs generated by LlaMA2-Chat}
    \label{fig:enter-thresholds-llamagen}

\end{figure}

Figure \ref{fig:thresholds-inputdata} shows the results of the representational analysis with the \textsc{McRae} and \textsc{ConceptNet} datasets. The dot products of the random pairs stand around zero, meaning that they are orthogonal to the \emph{partOf} concept direction. The distribution of the target pairs spans from negative to positive values.
Though they are decisively skewed towards the latter, more than $25\%$ of them have dot products below zero, as illustrated in Table \ref{tab:products-stats}. The number of target pairs whose dot products have higher values, and consequently a stronger alignment with the \emph{partOf} direction vector, is much lower: less than $50\%$ of pairs have dot products inferior to the average, and only between $10\%$ and $14\%$ of them have values above the maximum value reported for the random pairs. Figure \ref{fig:enter-thresholds-llamagen} and Table \ref{tab:products-stats-gendata} show that the same analysis conducted on the pairs generated by LlaMa2-7b-chat has a very similar trend, though with some improvements, like in the behavioral task of self-generated meronymy understanding (cf. Section \ref{self-gen-res}). Overall, the linear representation hypothesis is only partially confirmed for the \emph{part-whole} relation, with several pairs having a weak alignment along the concept direction. Additionally, we notice that there is no significant difference between the embedding and the unembedding spaces, as well as between the instruct and non-instruct models. 


\begin{table}
\begin{tabular}{lcc|clcc}
 & \multicolumn{2}{c|}{\textbf{\begin{tabular}[c]{@{}c@{}}Target Pairs\\ Dot Product\end{tabular}}} & \multicolumn{4}{c}{\textbf{\begin{tabular}[c]{@{}c@{}}Target Pairs\\ above the thresholds\end{tabular}}} \\ \hline
\multicolumn{1}{l|}{\textbf{Space}} & Avg & Std & \multicolumn{2}{c}{0} & Avg & Max RP \\ \hline
\multicolumn{1}{l|}{LlaMA2 Embeddings} & 7.63 & 12.36 & \multicolumn{2}{c}{72\%} & 45\% & 10\% \\
\multicolumn{1}{l|}{LlaMA2 Unembeddings} & 10.68 & 28.67 & \multicolumn{2}{c}{69\%} & 36\% & 13\% \\
\multicolumn{1}{l|}{LlaMA2-Chat Embeddings} & 7.60 & 12.13 & \multicolumn{2}{c}{72\%} & 45\% & 10\% \\
\multicolumn{1}{l|}{LlaMA2-Chat Unembeddings} & 8.63 & 13.34 & \multicolumn{2}{c}{73\%} & 42\% & 14\%\\
\hline
\end{tabular}%

\caption{Summary statistics of the representational analysis for the data derived from \textsc{CONCEPTNET} and \textsc{MCRAE}. RP stands for Random Pairs, whose dot products with $\hat{\gamma}(partOf)$ have an average value of 0 and a standard deviation of 1.41 across all spaces.}
\label{tab:products-stats}
\end{table}

\begin{table}[t]
\begin{tabular}{lccclcc}
 & \multicolumn{2}{c}{\textbf{\begin{tabular}[c]{@{}c@{}}Target Pairs\\  Dot Product\end{tabular}}} & \multicolumn{4}{c}{\textbf{\begin{tabular}[c]{@{}c@{}}Target Pairs\\  above the thresholds\end{tabular}}} \\ \hline
\multicolumn{1}{l|}{\textbf{Space}} & Avg & \multicolumn{1}{c|}{Std} & \multicolumn{2}{c}{0} & Avg & Max RP \\ \hline
\multicolumn{1}{l|}{LlaMA2 Embeddings} & 12.54 & \multicolumn{1}{c|}{15.07} & \multicolumn{2}{c}{81\%} & 53\% & 20\% \\
\multicolumn{1}{l|}{LlaMA2 Unembeddings} & 23.32 & \multicolumn{1}{c|}{34.30} & \multicolumn{2}{c}{80\%} & 26\% & 26\% \\
\multicolumn{1}{l|}{LlaMA2-Chat Embeddings} & 12.35 & \multicolumn{1}{c|}{14.66} & \multicolumn{2}{c}{81\%} & 54\% & 20\% \\
\multicolumn{1}{l|}{LlaMA2-Chat Unembeddings} & 13.94 & \multicolumn{1}{c|}{16.66} & \multicolumn{2}{c}{81\%} & 48\% & 28\%\\
\hline
\end{tabular}%

\caption{Summary statistics of the representational analysis for the data generated by LlaMA2-chat. RP stands for Random Pairs, whose dot products with $\hat{\gamma}(partOf)$ have an average value of 0 and a standard deviation of 1.41 across all spaces.}
\label{tab:products-stats-gendata}
\end{table}

The global analysis shows a great variability in the encoding of meronymy in vector spaces. This suggests that a model may encode linearly not the whole set of instances of the \textit{part-whole} relation list, but only some specific subsets of it. We explored this hypothesis by investigating the encoding of meronymic pairs belonging to four semantic classes: \emph{birds}, \emph{mammals}, \emph{houses/buildings}, and \emph{vehicles}. For each class, we chose a set of seed holonyms and then randomly selected a subset of meronymic pairs for each holonym (see Table \ref{tab:tp4class}). For each class $c$ we generated a specific $\hat{\gamma}(partOf_c)$ direction vector, and we performed a cross-comparison computing the alignment between every target pair and each direction vector. 


\begin{table}[t]
\begin{tabular}{l|cc}
Class & Seed Holonyms & Target Pairs \\
\hline
Vehicles & 7 & 51 \\
Mammals & 11 & 51 \\
Houses/Buildings & 10 & 39 \\
Birds & 17 & 51\\
\hline
\end{tabular}%
\caption{Number of seed holonyms and selected target pairs for each class.}
\label{tab:tp4class}
\end{table}

\begin{figure}
    \centering
    \includegraphics[width= \textwidth]{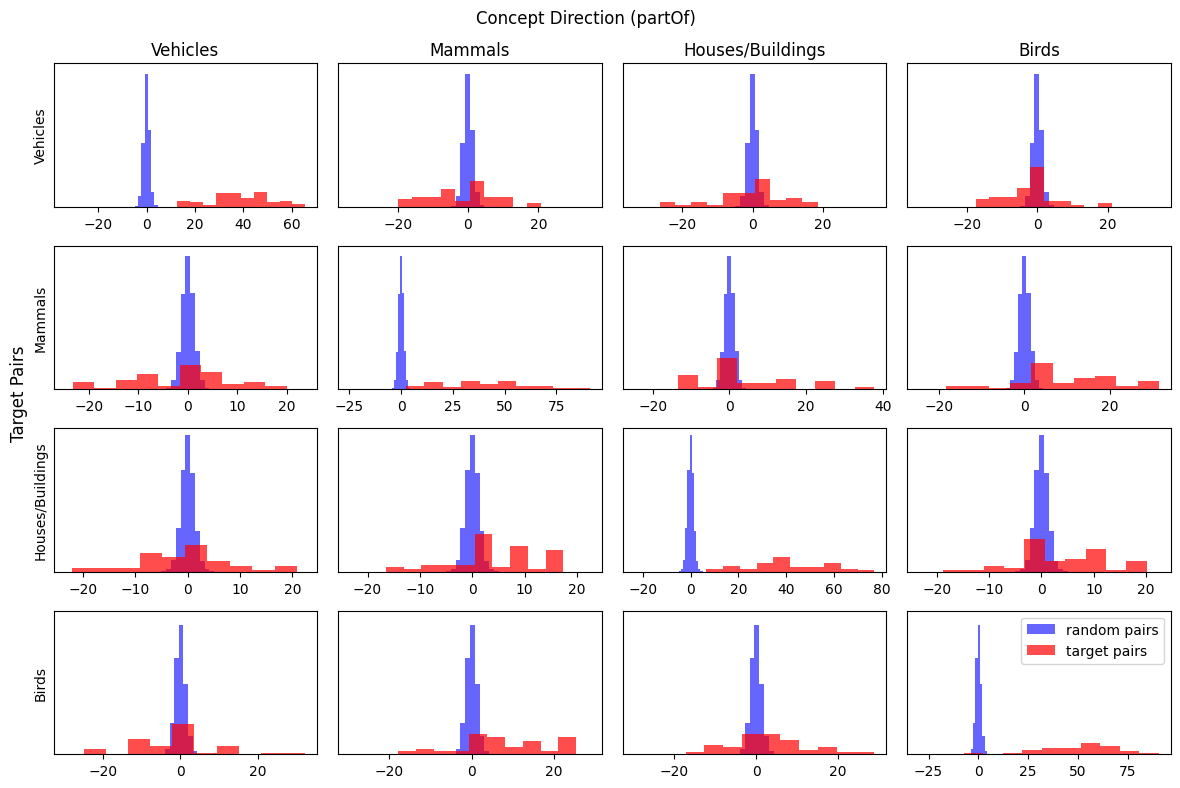}
    \caption{Cross comparison of the product distributions of class-specific meronymic target pairs with respect to random ones.}
    \label{fig:selected pairs}
\end{figure}

\begin{figure}
    \centering
    \includegraphics[width=0.8\textwidth]{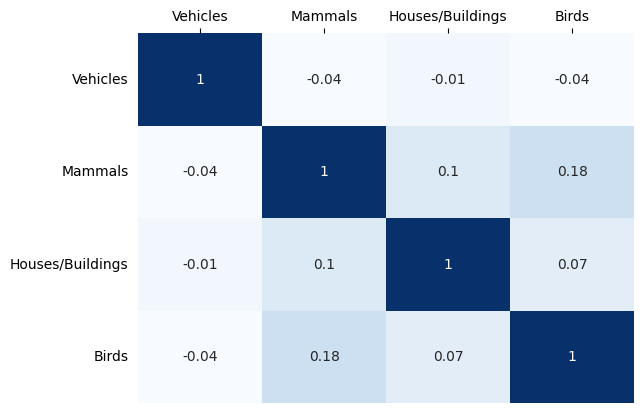}
    \caption{Cross comparison of the dot products between the class-specific $partOf_c$ direction vectors.}
    \label{fig:dir-align}
\end{figure}

Here, we report the analysis performed on the LlaMa2-7b-chat unembedding layer, but the other spaces show very similar trends. As can be seen in the diagonal of Figure  \ref{fig:selected pairs}, each class $c$ of target pairs is pretty aligned with the corresponding  $partOf_c$ concept direction, but it is not aligned with the direction vectors of the other classes, since their dot products are similar or even inferior to the ones of the random pairs. 
In fact, Table \ref{tab:cross-comp-stat} reveals that the average dot products between target pairs and direction vectors of their same class are significantly higher than those obtained with other classes. These values are also sensibly higher than those obtained in the global analysis (see Table \ref{tab:products-stats}).
This suggests that the embedding spaces linearly encode class-specific $partOf$ concepts, rather than a general meronymic relation. Thus, vector representations seem to fail to encode the abstract relation of meronymy, and distinct classes of \emph{part-whole} items correspond to very different directions in vector spaces.  This is further corroborated by the analysis of the direction vectors themselves. If the class-specific $partOf_c$ vectors were instances of a more general and abstract $partOf$ direction, we should expect them to be strongly aligned. As shown in Figure \ref{fig:dir-align}, this is not the case and the class-specific $partOf_c$ vectors point towards very different directions.

\begin{table}[]
\begin{tabular}{lllllllll}
 & \multicolumn{8}{c}{\textbf{Class-specific Direction Vectors}} \\ \cline{2-9} 
\textbf{} & \multicolumn{2}{l}{\textbf{Vehicles}} & \multicolumn{2}{l}{\textbf{Mammals}} & \multicolumn{2}{l}{\textbf{Houses/Bulidings}} & \multicolumn{2}{l}{\textbf{Birds}} \\ \hline
\multicolumn{1}{l|}{\textbf{Target Pairs}} & Avg & \multicolumn{1}{l|}{Std} & Avg & \multicolumn{1}{l|}{Std} & Avg & \multicolumn{1}{l|}{Std} & Avg & Std \\ \hline
\multicolumn{1}{l|}{\textbf{Vehicles}} & \textbf{33} & \multicolumn{1}{l|}{12} & -2 & \multicolumn{1}{l|}{9} & -1 & \multicolumn{1}{l|}{10} & -2 & 8 \\
\multicolumn{1}{l|}{\textbf{Mammals}} & -2 & \multicolumn{1}{l|}{10} & \textbf{34} & \multicolumn{1}{l|}{21} & 4 & \multicolumn{1}{l|}{12} & 9 & 11 \\
\multicolumn{1}{l|}{\textbf{Houses/Buildings}} & -1 & \multicolumn{1}{l|}{9} & 4 & \multicolumn{1}{l|}{8} & \textbf{32} & \multicolumn{1}{l|}{15} & -2 & 10 \\
\multicolumn{1}{l|}{\textbf{Birds}} & -2 & \multicolumn{1}{l|}{10} & 8 & \multicolumn{1}{l|}{11} & 3 & \multicolumn{1}{l|}{10} & \textbf{44} & 20\\
\hline
\end{tabular}%
\caption{Cross comparison of the average product and standard deviation between target pairs and different specifically identified concept directions.}
\label{tab:cross-comp-stat}
\end{table}

\section{General Discussion} \label{sec:wrap-up}
In this work, we have challenged LLMs on their understanding of the \textit{part-whole} relation, focusing on its core property of antisymmetry, as an aspect of the more general problem of assessing the true inferential semantic competence of such models \cite{Marconi2003-nu, lenci:2023b}. We did so through different experimental methodologies, trying to account for a variety of aspects going from behavioral responses to concept representations. However, the underlying questions of this inquiry, whether pursued with prompting, probability scoring or analysis of concept representation all regarded several aspects concerning the ability of LLMs to deal with parts, wholes and the relation holding between them. 


\paragraph{\textbf{Behavioral analysis}} We used prompt-based behavioral analysis to investigate the ability of three Large Language Models, namely LlaMa2-7b, LlaMa2-7b-chat and GPT-4, to satisfy the \textbf{Meronymy Knowledge Criterion} (MKC) defined in \ref{MKC}, which incorporates the assumption that real knowledge of the \emph{part-whole} relations entails knowledge of its antisymmetric nature (cf. in Section \ref{behav-task-desc}).
We tested the MKC in two tasks, \textbf{binary question answering} and \textbf{binary statement verification}, which we fed to the models to test their generalization abilities across different ways to probe the same type of knowledge. 
GPT4 showed the strongest knowledge of \textit{meronymy} and the highest accuracy in passing the MKC. However, even this top model reaches a maximum of $69\%$ accuracy on the \textsc{conceptnet} dataset (see Figure \ref{fig:task1-global}), with the performance of the other models being much lower.

It has to be noted that all three LLMs seem to struggle more with \textsc{conceptnet}.
This might be due to greater level of specificity and technicality of the \textsc{conceptnet} meronyms.
On the other hand, data derived from \citet{McRae:2005} are essentially judgments given by people when asked to list the features of a given physical object, without the specific aim of listing meronyms, which may have led people to express more general, superficial, and yet distinctive parts for the target entities. 

We did not observe a substantial difference between the instruct and non-instruct versions of LlaMA-7b when asking the model to answer questions.
A comparison between the binary question answering and the binary statement verification tasks instead reveals a drop in average performance when moving from the former to the latter. As they are conceptually the exact same task framed differently at the linguistic level, this discrepancy suggests that the LlaMA-7b models do not possess a real abstract knowledge about the \textit{part-whole} relation and its properties, failing to generalize in different contexts and under different linguistic formulations. A different case has to be made for GPT4, whose performances are very similar between \texttt{questions} and \texttt{statements} with a very slight advantage for the former. This can be traced directly to the greatest overall capabilities of GPT-4, partially due to its superior size and allegedly better alignment training. However, the performance gap shown by GPT-4 also reveals the imperfection of its generalization ability. 

The second behavioral task consisted in the \textbf{generation of parts} by the instruct models. The results show their remarkable ability to generate parts for objects given as input (see Table \ref{tab:partgen}). Indeed, LlaMA2-7b-chat generated twice as much meronyms as those in the \textsc{conceptnet} and \textsc{mcrae} datasets, and GPT4 almost ten thousand more. Model-generated parts tend to be more diverse and sometimes specific than human-generated ones. This is consistent with previous work on property generations: \citet{Hansen2022} found that asking GPT-3 to generate feature norms for objects resulted in a more varied and detailed set of norms compared to human-generated ones.

The items for which the LLMs generated more parts usually refer to vehicles and artifacts, and mechanical devices (i.e., \textit{ship}, \emph{aeroplane}, \emph{car}, etc.). A significant number of parts was also listed for animals.
The generation task seems to be the one at which the models are most proficient, though it is hard to quantitatively evaluate it, because of the lack of a reliable gold standard in an open-ended generation task to automatically assess the models' performance. Moreover, this is the task in which the theoretical concerns about meronymy and its conceptual-linguistic ambiguities (cf. Section \ref{part-whole-rel}) may become most impactful. 
The intrinsic semantic ambiguity of the term "\textit{part}", as outlined briefly above in Section \ref{part-whole-rel}, makes it difficult to evaluate whether proper parts have been listed for certain entities.

However, we carried out a qualitative analysis of some of the most problematic models' generations. One common case is represented by entities that have several optional parts, which are not essential to their functioning nor to their ontological definition but may be typical or frequent. For instance, the meronym \emph{keyboard drawer} is not essential in the description of a holonym like \emph{desk}, though indeed desks come frequently with keyboard drawers. Models also generate both \textit{systemic} and \textit{segmental} parts (cf. Section \ref{part-whole-rel}). For example, \textit{air conditioning} for \textit{house} or \textit{skews} for \textit{table} should be considered as \textit{systemic} parts, while \textit{room} or \textit{legs} segmental ones \citep{cruse-lexsem}.

Another interesting, albeit rare, case concerns the generation of a particular instance of the given holonym rather than a real meronym. For example, for the word \textit{temple}, LLMs generated names actual temples (i.e., \textit{haghia sofia, Luxor}). It is interesting to note that some studies have shown how confusion between \textit{taxonomy} and \textit{part-whole} relations may take place in human subjects \citep{parton-taxon-conf}, which contribute to fueling the ambiguous theoretical status of the \textit{part-whole} relation. 
These cases may be due to the ambiguity of the term \emph{part} itself and to the out-of-context usage of the target holonyms in the prompt. For instance, temples may vary greatly across cultures and contexts, which makes it difficult to describe it, and hence to enumerate its parts, without further specification (e.g., \emph{Roman temple} vs. \emph{Buddhist temple}). 

The last behavioral task tested whether the the models satisfied the MKC on their own generated parts, a task we called \textbf{self-generated meronymy understanding}. This allowed us to check how performance changes when scaling up the input data (given the much greater number of test meronymic pairs) and when using models' own generated data \citep{west2023generative}.
We observed quite the same performance as with the \textsc{conceptnet} and \textsc{mcrae} datasets used in the first task, though having sharply scaled up the number of inputs. In several cases, models do not correctly understand the very same parts they had generated (see Figure \ref{fig: res_from_gen}). This is consistent  with what \citet{west2023generative} dubbed the \textbf{Generative AI Paradox}, according to which generative AI models are far better at generating than at understanding, to the point that sometimes they may not even understand their own outputs. This is particularly true for LlaMA2-7b-chat, which got the worse performance on its own generated data, not even reaching the chance level when asked to judge \texttt{statements} derived from its own generated meronyms and the respective swapped versions. 

Overall, the behavioral analysis has shown a relative instability in the capabilities of the tested LLMs of abstracting and generaliing robustly the meronymy relation, and has revealed still quite a limited understanding of its core inferential property of \textit{antisymmetry}.



\paragraph{\textbf{Probabilistic analysis}} We compared the probability scores assigned by the LLMs to pairs of \textbf{plausible and implausible statements}, with the former being real examples of \textit{part-whole} relation (e.g., \emph{A wheel is a part of a car}), and the latter counterfactual sentences  violating the \textit{anti-symmetry} property of meronymy (e.g., \emph{A car is a part of a wheel}). This probabilistic analysis complements the results obtained in the behavioral one, smoothing the limitations posed by prompt-based tasks. In fact, while a given prompt may not be the perfect fit to lead the model to output the desired answers, causing models' responses to be unstable and making it difficult to correctly estimate their performance, probability scores may be more solid and a better indicator of the linguistic abilities of LLMs \citep{hu-levy-2023-prompting}. In tasks involving meta-linguistic (and meta-cognitive) requests formulated with prompting, models do not only have to retrieve the relevant knowledge from their internal parameters, but they also have to do so in a specific manner, demonstrating the ability to follow the instructions given by humans. This setting makes the execution of the task noisier. On the contrary, feeding the models with simple sentences and extracting the probabilities they assign to them, may be a faster and more genuine way of assessing the abilities of these models to represent linguistic and cognitive phenomena in probability space.
 
The two models tested on this task -- LlaMa2-7b and LlaMa2-7b-chat -- perform soundly above the chance level on the dataset containing oppositions of veridical and counterfactual instances,  while sticking around random guessing when moving to the set of fake meronyms pairs (see Section \ref{fig:sents-prob}). This is a hint of the models' ability to discriminate between semantically plausible meronymic relations and implausible ones. This perfomance is higher than the one obtained by the same models in satisfying the MKC in the prompt-based tasks. However, a conspicuous gap still remains, with an error rate attested around $25\%$, representing cases in which plausible and implausible meronymic sentences are not properly distinguished in the models' probability space. 



\paragraph{\textbf{Representational Analysis}} In Section \ref{sec:conc}, we investigated how the \textit{part-whole} concept is represented in the LLMs' embedding space. We followed the \textbf{linear representation hypothesis} and applied the methodology by \citet{park2024linearrepresentationhypothesisgeometry} to search for some cue of linearity in the structure of the encoding of \textit{part-whole} representative pairs in the embedding and unembedding spaces of both LlaMa2-7b and LlaMa2-7b-chat models. 
We found that indeed \textit{some hints} of linear structure for meronymy encoding are recoverable from the embedding and the unembedding spaces,
but more coherent linearity can only be observed for class-specific meronyms, such as ``parts of vehicles'' or ``parts of mammals''.  Conversely, meronyms belonging to different semantic categories appear to correspond to orthogonal directions in the embedding space. This suggests that the models lack a general abstract representation of the \textit{part-whole} concept, but they represent semantically coherent sub-categorization of this relation.
Recalling the problems posed by the theoretical status of the \textit{part-whole} relation and its alleged internal structure in terms of taxonomical organisation (Section \ref{part-whole-rel}), we may indeed expect to find more linear structures when we move to manually selecting and grouping representative pairs of \textit{specific part-whole} relation.
On the other hand, it seems that at least for meronymy, its encoding in the models internal representations is mostly driven by shared similarities between the corresponding holonyms (e.g., mammals, vehicles, etc.), rather than by a truly general representation of the of \emph{part-whole} concept. This might be a hint of possible misalignments between semantic relation encoding in LLMs and in humans.
The development of a methodology to automatically discover sub-clusters of concepts which exhibit linear structure in vector space is be a promising avenue that we leave for future research.


\section{Conclusions} \label{sec:conc}

The success and appearance of more and more powerful LLMs have sparkled an intense debate on the real depth of their semantic knowledge. While some have emphasized the intrinsic impossibility for such models to acquire full semantic competence because of their lack of grounding on the external world \cite{bender-koller-2020-climbing}, others have instead stressed the fact that distributional statistics extracted from the huge training corpora would allow them to acquire competence at least inferential dimensions of meaning \cite{piantadosi2022meaning,Sgaard2022,Pavlick2023}. 
In this work, we investigated one aspect of such inferential competence. We tested three LLMs on their knowledge of the \textit{part-whole} relation, their ability to correctly deal with its inferential property of \textit{antisymmetry} and their way of representing such concept in \textit{embedding}
space.
Given the complexity of the \textit{part-whole} relation, both theoretically and empirically, future work is reserved to explore the \textit{transitivity} property of \textit{meronymy} as well as further specifications of this relation, such as its possible subtypes.

The three methods we adopted for our investigation provide different and yet complementary perspectives, which nonetheless offer quite a coherent view,  when combined together. Although simple and pretty straightforward, the prompting tasks based on binary question answering and truth statements judgment showed that the models exhibit neat limitations in abstracting the \textit{part-whole} relation, and indeed fail to consistently give correct responses with respect to the antisymmetry property of meronymy. 
However, prompt-based evaluation might underestimate the models' ability, overlooking latent knowledge encoded into the models, which would fail to surface when elicited through behavioral testing. While prompting is the most natural diagnostic for querying LLMs and probing their knowledge, it has the drawback of being unstable and pretty shallow. It is fairly understood that minimal changes in the prompt configuration may be reflected in massive fluctuations in models' performances \citep{zhang-etal-2023-beyond, sclar2023prompt-sensitivity}. A further proof for this is given by the different models' accuracy in satisfying the MKC, whether this is formulated as a statement verification task, or a question answering one. Failure to output certain knowledge cannot be taken as a decisive clue of the lack of encoding of such knowledge in the model \citep{burns2024discovering}. As suggested by recent findings in the literature \citep{hu-levy-2023-prompting, kauf_etal:2023, kauf-etal-2024-log}, probability scores give us a measure of a system's ability to model a certain probability distribution, hence discriminating semantically plausible sentences from implausible ones. However, despite improving over prompt-based tasks, probabilistic analysis also shows that models have only a limited knowledge of the \textit{part-whole} relation. 
The results of representational analysis, point towards a similar conclusion from a concept representation standpoint. However, linking concept encoding in LLMs' representations to their behavioral performance is not straightforward. Additionally, the linear representation hypothesis has still an unclear empirical status, and its acceptability is still debated and controversial \citep{engels2024languagemodelfeatureslinear, csordasrecurrent, strong-weak-lrh}. However, results show just a weak encoding of the \emph{part-whole} relation in embeddings spaces, mostly restricted to semantically similar items.

Overall, these analyses suggest that LLMs have at best only a \textbf{partial} mastery of meronyny and its inferential properties. Though impressive it may look \emph{prima facie}, it is just a ``\emph{quasi}-semantic'' competence, with still significant differences from the human one, a gap that is even striking  given the huge amounts of linguistic data the models are trained on. Though these data allow models to generate a lot of parts of objects with a very high accuracy, they do not seem to be sufficient to grant them a similarly high accuracy in understanding meronymy, even when they are tested on their self-generated parts. This might suggest that the \emph{part-whole} relation is only partly recoverable from textual data, consistently with the embodied nature of such relation, which is deeply grounded in our experience in the world \cite{Croft:Cruse:2004}. It might also indicate that distributional statistics, such as the one LLMs rely on, can only approximate a shallow notion of meronymy, but are not enough to acquire deeper inferential properties.



\starttwocolumn
\bibliography{compling_style}

\end{document}